\documentclass[sigconf, screen]{acmart}
\renewcommand\footnotetextcopyrightpermission[1]{} % removes footnote with conference information in first column
\usepackage{url}
\usepackage{xcolor}
\definecolor{newcolor}{rgb}{.8,.349,.1}
\usepackage{enumitem}
\usepackage{enumerate}
\usepackage{colortbl}
\usepackage{float}
\usepackage[ruled]{algorithm2e}
\usepackage{subcaption}
\usepackage{multirow}
\definecolor{mygray}{gray}{.9}
\definecolor{ForestGreen}{RGB}{34,139,34}
\newcommand\ie{i.e.}

\usepackage{tcolorbox}
\newtcolorbox[list inside=prompt,auto counter,number within=section]{prompt}[1][]{
    colbacktitle=black!60,
    coltitle=white,
    fontupper=\footnotesize,
    boxsep=5pt,
    left=0pt,
    right=0pt,
    top=0pt,
    bottom=0pt,
    boxrule=1pt,
    title={#1},
    #1, % add more args
}
\AtBeginDocument{%
  }

\newcommand\dataset{AV-Deepfake1M++}
\newcommand\datasetabbr{AV-Deepfake1M++}

\setcopyright{acmlicensed}
\copyrightyear{2025}
\acmYear{2025}
\acmDOI{XXXXXXX.XXXXXXX}
\acmConference[ACMMM '25]{33rd ACM International Conference on Multimedia}{October 27--31, 2025}{Dublin, Ireland}
\begin{document}

\title[AV-Deepfake1M++]{AV-Deepfake1M++: A Large-Scale Audio-Visual Deepfake Benchmark with Real-World Perturbations}


\author{Zhixi Cai}
\email{zhixi.cai@monash.edu}
\orcid{0000-0001-7978-0860}
\affiliation{%
 \institution{Monash University}
 \city{Melbourne}
 % \state{Western Australia}
 \country{Australia}}

\author{Kartik Kuckreja}
\email{kartik.kuckreja@mbzuai.ac.ae}
\orcid{}
\affiliation{%
  \institution{MBZUAI}
 \city{Abu Dhabi}
 % \state{Western Australia}
 \country{United Arab Emirates}
}

\author{Shreya Ghosh}
\email{shreya.ghosh@curtin.edu.au}
\orcid{0000-0001-8356-4909}
\affiliation{%
  \institution{Curtin University}
 \city{Perth}
 % \state{Western Australia}
 \country{Australia}
}
\author{Akanksha Chuchra}
\email{akanksha.22csz0001@iitrpr.ac.in}
\orcid{}
\affiliation{%
  \institution{IIT Ropar}
 \city{Ropar}
 % \state{Western Australia}
 \country{India}
}

\author{Muhammad Haris Khan}
\email{muhammad.haris@mbzuai.ac.ae}
\orcid{0000-0001-9746-276X}
\affiliation{%
  \institution{MBZUAI}
  \city{Abu Dhabi}
  % \state{Victoria}
  \country{United Arab Emirates}
}

\author{Usman Tariq}
\email{utariq@aus.edu}
\orcid{0000-0002-8244-2165}
\affiliation{%
  \institution{American University of Sharjah}
  \city{Sharjah}
  % \state{Victoria}
  \country{United Arab Emirates}
}

\author{Tom Gedeon}
\email{tom.gedeon@curtin.edu.au}
\orcid{0000-0001-8356-4909}
\affiliation{%
  \institution{Curtin University}
 \city{Perth}
 % \state{Western Australia}
 \country{Australia}
}

\author{Abhinav Dhall}
\email{abhinav.dhall@monash.edu}
\orcid{0000-0002-2230-1440}
\affiliation{%
  \institution{Monash University}
  \city{Melbourne}
  % \state{Victoria}
  \country{Australia}
}

\renewcommand{\shortauthors}{Cai et al.}

\begin{abstract}
 
The rapid surge of text-to-speech and face-voice reenactment models makes video fabrication easier and highly realistic. To encounter this problem, we require datasets that rich in type of generation methods and perturbation strategy which is usually common for online videos. To this end, we propose AV-Deepfake1M++, an extension of the AV-Deepfake1M having 2 million video clips with diversified manipulation strategy and audio-visual perturbation. This paper includes the description of data generation strategies along with benchmarking of AV-Deepfake1M++ using state-of-the-art methods. We believe that this dataset will play a pivotal role in facilitating research in Deepfake domain. Based on this dataset, we host the 2025 1M-Deepfakes Detection Challenge. The challenge details, dataset and evaluation scripts are available online under a research-only license at \url{https://deepfakes1m.github.io/2025}.

\end{abstract}


% \ccsdesc[500]{Computing methodologies~Computer vision}
% \ccsdesc[500]{Security and privacy~Social aspects of security and privacy}
% \ccsdesc[300]{Security and privacy~Usability in security and privacy}

% A "teaser" image appears between the author and affiliation
% information and the body of the document, and typically spans the
% page.


\maketitle

\section{Introduction}
\label{sec:intro}
In this era of generative AI, highly realistic audio, visual contents blurs the gap between real and fake contents, even for humans~\cite{zhouFace2021, narayanDFPlatter2023,cai1MDeepfakes2024}. This growing ambiguity creates opportunities to utilize the malicious use of GenAI technologies, including the spread of misinformation. To address this issue, the development of robust and reliable detection methods has become critically important. The quality of deepfake detector heavily relies on large, diverse benchmarking efforts to drive progress in both coarse-grained deepfake detection and fine-grained temporal localization.  

Benchmarking effort in deepfake domain have been evolved from face-swap imagery (i.e. FaceForensics++~\cite{zhouFace2021} and DFDC~\cite{dolhanskyDeepFake2020}) to cross-modal manipulations and word-level edits such as the FakeAVCeleb~\cite{khalidFakeAVCeleb2021} and the AV-Deepfake1M~\cite{cai1MDeepfakes2024}. A holistic overview of deepfake datasets is shown in \autoref{tab:datasets}. Despite the current progress in benchmarking effort, three key gaps remain mentioned below:
\textit{First of all}, scale and source diversity remain limited in AV-Deepfake1M as it relies solely on VoxCeleb2~\cite{chungVoxCeleb22018}. VoxCeleb2~\cite{chungVoxCeleb22018} was curated for single speaker situation which restrict the demographic coverage and real-world linguistic richness.
\textit{Secondly}, the benchmarks mentioned in \autoref{tab:datasets} lacks in terms of generation diversity. The deepfake benchmarks need to keep pace with the explosion of synthesis techniques. Most prior benchmarks rely on one visual and at most two text-to-speech back-ends. Less diversity on training data encourage overfitting to specific artifacts.
\textit{Thirdly}, in prior benchmarking AV-Deepfake1M~\cite{caiAVDeepfake1M2024}, streaming and redistribution artifacts such as blur, re-compression, frame drops, reverberation, packet jitters are overlooked, despite being ubiquitous in real-world scenarios. These artifacts can obscure forensic cues or may introduce misleading signals, thereby complicating the detection of forgeries.

To overcome the aforementioned issues, we propose a new benchmark, \datasetabbr{}, for audio-visual deepfake detection and localization tasks. The main contribution of this paper is as follows: 
%------------------------------------------------------------------------------
\begin{itemize}[leftmargin=1.2em, itemsep=2pt]
\item To the best of our knowledge, \datasetabbr{} is the large scale and diverse dataset containing 2 million clips (\(\sim\)4\,600\,h) curated from three different source datasets; VoxCeleb2, LRS3, and EngageNet. \datasetabbr{} contains diverse situations like studio interviews, TED talks, and natural conversational meetings.

\item The deepfake generation pipeline includes nine state-of-the-art models such as \textit{visual-LipSync}, \textit{LatentSync}, \textit{Diff2Lip}; \textit{audio-VITS}, \textit{YourTTS}, \textit{F5TTS}, \textit{XTTSv2}, \textit{VALLEX}. We incorporate these models to create unimodal as well as cross-modal forgeries with \textit{insert}, \textit{replace} and \textit{delete} strategies.

\item To address the real-world perturbations including streaming and redistribution artifacts, we integrate 15 video-level and 11 audio-level distortions such as Gaussian/Poisson noise, rolling-shutter, colour quantisation, Doppler shift, clipping, \emph{etc.}. \datasetabbr{} have a held-out test set which further adds difficulty level with mixed perturbation schedules such as frame-rate jitter or audio stutter.

% \item \datasetabbr{} have extensive benchmarking with 8 visual, 4 audio, 3 multimodal detectors, 2 temporal localization frameworks. There is a substantial degradation in performance as compared with results on \textit{AV-Deepfake1M} and \textit{LAV-DF}.
\end{itemize}
%------------------------------------------------------------------------------
% Taken together, these upgrades transform our previous dataset into the \emph{largest and most challenging} public resource for audio-visual deepfake research.  
% Initial experiments show that even the strongest temporal forgery localiser, UMMAFormer, loses over one-third of its AP when confronted with AV-Deepfake1M{\small++}, while frame-based Xception drops more than 15\,AUC points.  
% We hope this benchmark will catalyse research into perturbation-robust representations, cross-modal consistency modelling and real-time defences deployable in the open Internet.

\begin{table}[t]
\centering
\caption{\textbf{Details for publicly available deepfake datasets in a chronologically ascending order.} \textmd{Cla: Binary classification, SL: Spatial localization, TL: Temporal localization, FS: Face swapping, RE: Face reenactment, TTS: Text-to-speech, VC: Voice conversion.}}
\label{tab:datasets}
\scalebox{0.68}{
\begin{tabular}{l||c|c|c|c|c|c}
\toprule[0.4mm]
\rowcolor{mygray}\textbf{Dataset} & \textbf{Year} & \textbf{Tasks} & \multicolumn{3}{c|}{\textbf{Manipulation}} & \textbf{\#Total} \\
\rowcolor{mygray}&  &  & \textbf{Mod.} & \textbf{Method} & \textbf{Source} &\\
\hline\hline
DF-TIMIT~\cite{korshunovDeepFakes2018} & 2018 & Cla & V & FS & - & 960 \\
UADFV~\cite{yangExposing2019} & 2019 & Cla & V & FS & -  & 98  \\
FaceForensics++~\cite{rosslerFaceForensics2019} & 2019 & Cla & V & FS/RE & -  & 5,000 \\
Google DFD~\cite{nickContributing2019} & 2019 & Cla & V & FS & - & 3,431 \\
DFDC~\cite{dolhanskyDeepFake2020} & 2020 & Cla & AV & FS & - & 128,154 \\
DeeperForensics~\cite{jiangDeeperForensics12020} & 2020 & Cla & V & FS &  - & 60,000 \\
Celeb-DF~\cite{liCelebDF2020} & 2020 & Cla & V & FS & -  & 6,229 \\
WildDeepfake~\cite{ziWildDeepfake2020} & 2020 & Cla & - & - &  - & 7,314 \\
FFIW$_{10K}$~\cite{zhouFace2021} & 2021 & Cla/SL & V & FS & -  & 20,000 \\
KoDF~\cite{kwonKoDF2021} & 2021 & Cla & V & FS/RE & -  & 237,942 \\
FakeAVCeleb~\cite{khalidFakeAVCeleb2021} & 2021 & Cla & AV & RE &  - & 25,500$+$ \\
ForgeryNet~\cite{heForgeryNet2021} & 2021 & SL/TL/Cla & V & Random FS/RE & - & 221,247 \\
ASVSpoof2021DF~\cite{liuASVspoof2023} & 2021 & Cla & A & TTS/VC & - & 593,253 \\
LAV-DF~\cite{caiYou2022} & 2022 & TL/Cla & AV & Content & - & 136,304 \\
% ADD 2022~\cite{yiADD2022} & 2022 & Cla & A & TTS/VC & 90 & 5,619 & 47,958 & 53.577~\footnote{The test set are not included.} \\
DF-Platter~\cite{narayanDFPlatter2023} & 2023 & Cla & V & FS & - & 265,756 \\
AV-Deepfake1M~\cite{caiAVDeepfake1M2024} & 2023 & TL/Cla & AV & Content &  LLM & 1,146,760 \\
% OHImg~\cite{may2023comprehensive}& 2023  & Diffusion & Overhead  &  13,150\\
% ArtiFact~\cite{rahman2023artifact} & 2023  & Diffusion \& GAN & Objects  & 2,496,738 \\
% AIGCD~\cite{zhong2023rich} & 2023  & GAN & Objects & 1,440,000 \\
% GenImage~\cite{zhu2023genimage} & 2023  & Diffusion \& GAN & Objects & 2,681,167 \\
M3Dsynth~\cite{zingariniM3DSYNTH2024}& 2024 & Cla & Img  & Diffusion & - & 8,577 \\
SemiTruths~\cite{palSemiTruths2024} & 2024 & Cla & Img & Diffusion & - & 1,500,300 \\
SIDA~\cite{huangSIDA2025} & 2024  & Cla & Img & Diffusion & - & 300,000\\
PolyGlotFake~\cite{houPolyGlotFake2025} & 2024 & Cla &  AV  & RE/TTS/VC & - & 15,238\\
Illusion~\cite{thakralILLUSION2024} & 2024 & Cla &  AV  & FS/RE/TTS & - & 1,376,371\\
MultiFakeVerse~\cite{guptaMultiverse2025} & 2025  & Cla &  Img  & VLM & VLM & 845,286\\
ArEnAV~\cite{kuckrejaTell2025}& 2025  & TL/Cla &  AV  & Content & LLM & 387,072\\ \hline
\dataset{} & 2025 & TL/Cla & AV & Content & LLM & \textbf{2,051,154} \\
\bottomrule[0.4mm]
\end{tabular}}
\vspace{-6mm}
\end{table}

\section{Related Work}

\noindent\textbf{Deepfake Datasets.} The earliest datasets targeted isolated visual manipulations such as face swapping or reenactment. DF-TIMIT~\cite{korshunovDeepFakes2018}, UADFV~\cite{yangExposing2019}, FaceForensics++ (FF++)~\cite{rosslerFaceForensics2019}, Google DFD~\cite{nickContributing2019}, DFDC~\cite{dolhanskyDeepFake2020}, KoDF~\cite{kwonKoDF2021} and DF-Platter~\cite{narayanDFPlatter2023} are all designed for coarse-grained binary classification task. ForgeryNet~\cite{heForgeryNet2021} and FFIW$_{10\text{k}}$~\cite{zhouFace2021} introduce spatial localization for deepfakes. There are further works focusing on the other aspects to enrich the deepfake diversity~\cite{kuckrejaTell2025, huangSIDA2025, guptaMultiverse2025}. However, these datasets only consider the deepfakes in single visual modality.

\begin{figure}[b]
\centering
\includegraphics[width=\linewidth]{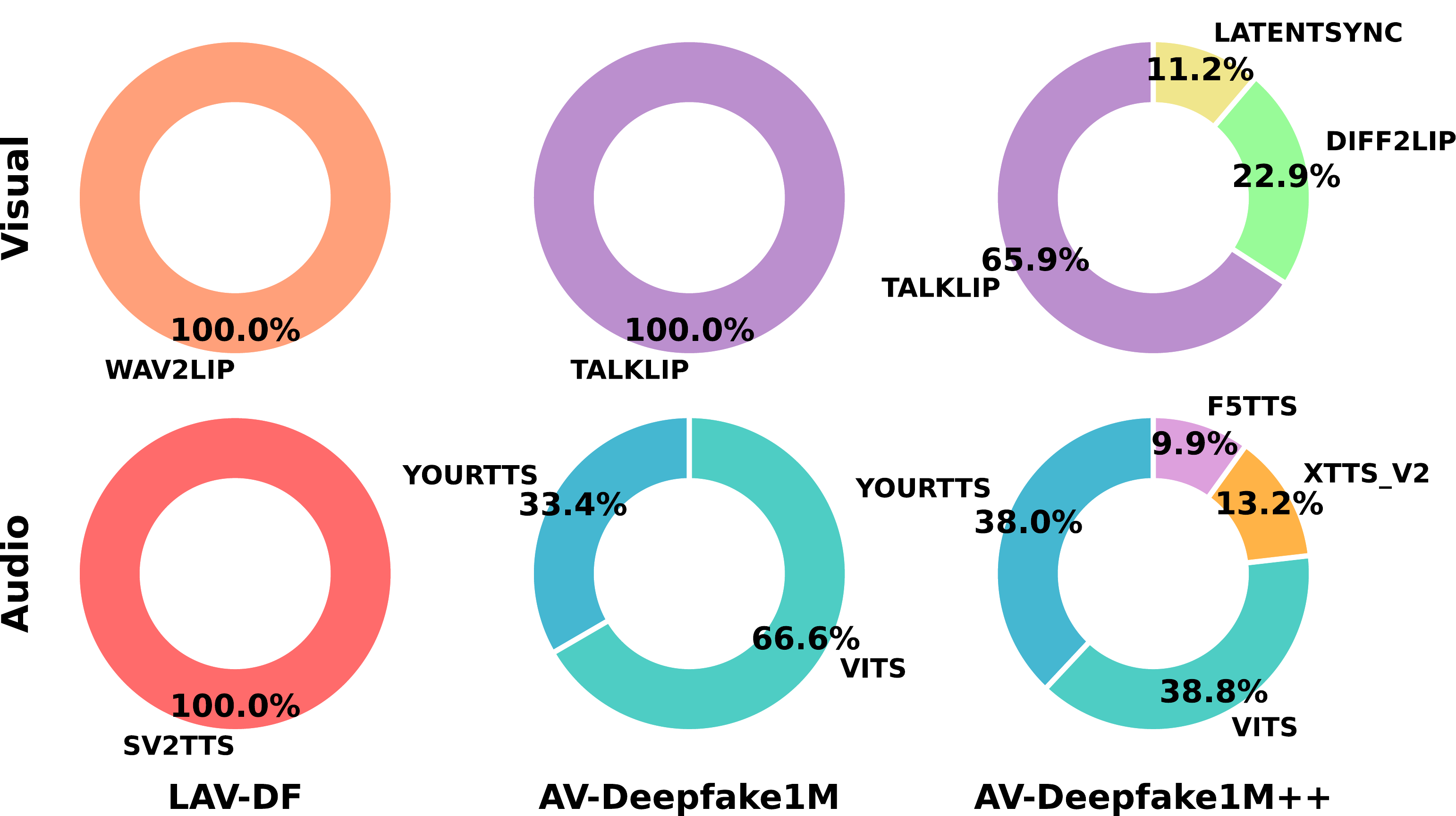}
\caption{\textbf{Comparison of LAV-DF, AV-Deepfake1M and \datasetabbr{} for deepfake generation methods.} \textmd{The first row shows the proportion of visual deepfake generation methods and the second row shows the proportions of the audio deepfake generation methods.}}
\label{fig:dataset_comparison}
\end{figure}

FakeAVCeleb~\cite{khalidFakeAVCeleb2021} extends the deepfake benchmark to the audio-visual multiple modalities. LAV-DF~\cite{caiYou2022} introduce a meaningful multimodal content-driven deepfakes that editing the word in the transcript to manipulate the video's content. However, both datasets rely on a \emph{single} visual generator (Wav2Lip~\cite{prajwalLip2020}) and a \emph{single} audio generator (SV2TTS~\cite{jiaTransfer2018}), encouraging detectors to overfit to method-specific artifacts. The rule-based text manipulation in LAV-DF limits the diversity of the generated deepfake content. AV-Deepfake1M~\cite{caiAVDeepfake1M2024} uses ChatGPT~\cite{ouyangTraining2022} and multiple higher-quality generators~\cite{casanovaYourTTS2022, wangSeeing2023, kimConditional2021} to improve the quality of the generated content, which is hard to be recognized by human. However, the real videos are solely from VoxCeleb2~\cite{chungVoxCeleb22018}, limiting the video diversity. Neither dataset models real-world redistribution artifacts (\ie compression, frame drop) that can suppress or mimic forensic cues. The previous datasets are shown in \autoref{fig:dataset_comparison}.

\noindent\textbf{Deepfake Generation.} Recent advances have dramatically lowered the barrier to high-fidelity deepfake generation. On the \textbf{visual} side, state-of-the-art (SoTA) lip-sync models such as LatentSync~\cite{liLatentSync2025}, Diff2Lip~\cite{mukhopadhyayDiff2Lip2024} and TalkLip~\cite{wangSeeing2023} outperform well-known predecessors~\cite{prajwalLip2020} in visual quality and temporal consistency. For SoTA audio zero-shot TTS methods (\ie XTTSv2~\cite{casanovaXTTS2024}, F5TTS~\cite{chenF5TTS2025}) clone a speaker’s voice from seconds of reference audio, while controllable prosody models can match emotion and style. The recent evolution of Large language models (LLMs) delivers the lower-cost, more efficient and better output quality LLMs, including GPT-4o~mini~\cite{openaiGPT4o2024a}, which can be used for automate semantic editing. LLM plans insert/replace/delete operations that keep syntax fluent yet invert meaning, a strategy already exploited in AV-Deepfake1M~\cite{caiAVDeepfake1M2024}. 

\datasetabbr{} bridges the gaps from previous works by (1) sourcing more real data from multiple datasets~\cite{chungVoxCeleb22018, afourasLRS3TED2018, singhHave2023}; (2) integrating more SoTA lip-sync~\cite{liLatentSync2025, mukhopadhyayDiff2Lip2024} and TTS methods~\cite{casanovaXTTS2024, chenF5TTS2025}; (3) simulating 36 audio/visual real-world perturbation; (4) providing frame-, and video-level annotations for both classification and temporal localization.

\begin{figure*}[t]
\centering
\includegraphics[width=0.9\textwidth]{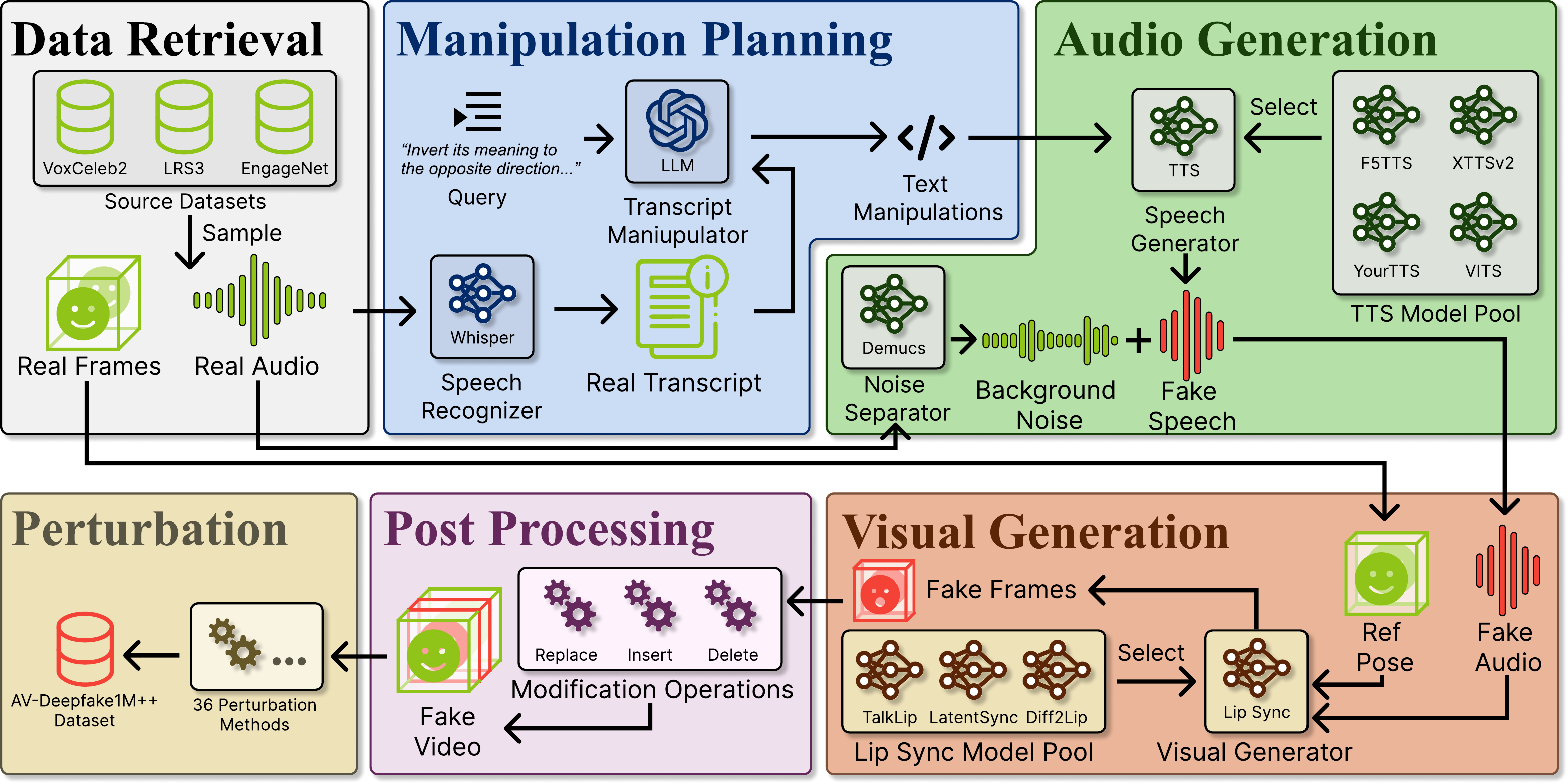}
\caption{\textbf{Data generation pipeline of \datasetabbr{}.} \textmd{}}
\label{fig:pipeline}
\end{figure*}

% -------------------------------------------------------------------------
\section{Dataset Generation}
\label{sec:dataset}

Figure~\ref{fig:pipeline} shows the pipeline we used to
build \datasetabbr{}. The pipeline inherits the structure of AV-Deepfake1M~\cite{caiAVDeepfake1M2024} but adds new source dataset, generation methods and perturbations.

\subsection{Data Retrieval}
\label{ssec:data}

We source unmanipulated original videos from three complementary datasets: VoxCeleb2~\cite{chungVoxCeleb22018}, LRS3~\cite{afourasLRS3TED2018}, and EngageNet~\cite{singhHave2023}. In the test subsets, videos are encoded with various codecs.

\subsection{Forgery Generation}
\label{ssec:fake}

\noindent\textbf{Manipulation Planning.} For every ASR transcript we invoke an LLM (GPT‑4o mini~\cite{openaiGPT4o2024a} and GPT-3.5 turbo~\cite{ouyangTraining2022}). The LLM receives a few‑shot prompt that asks it to invert the semantic
stance of the utterance in several token‑level operations. Operations are chosen from \emph{replace}, \emph{delete} and \emph{insert} and returned in a JSON schema
\emph{\{operation, old\_word, new\_word, index\}}. 

\noindent\textbf{Audio Generation.}
We separate speech and background noise with Demucs~\cite{defossezReal2020}. Text‑to‑speech (TTS) synthesis then produces manipulated speech in two paradigms: few-shot method VITS~\cite{kimConditional2021}, zero-shot methods F5TTS~\cite{chenF5TTS2025}, XTTSv2~\cite{casanovaXTTS2024} and YourTTS~\cite{casanovaYourTTS2022}. For \emph{replace/insert} we generate either (i) the whole modified sentence and crop the required span, or (ii) only the new word(s), yielding two slightly different pipeline. For \emph{delete} we keep only background noise. All outputs are loudness‑matched to the original audio.

\noindent\textbf{Visual Generation.}
Audio-driven lip‑sync frames are synthesized with a model pool: TalkLip~\cite{wangSeeing2023}, LatentSync~\cite{liLatentSync2025}, and Diff2Lip~\cite{mukhopadhyayDiff2Lip2024}.
The reference head pose is sampled from the position to be manipulated.

\noindent\textbf{Post Processing.}
Depending on the manipulation plan, the generated
\emph{replace}, \emph{insert} or \emph{delete} segments are assembled into the real video. We also follow the previous dataset~\cite{caiAVDeepfake1M2024} generating 4 types of the manipulations: \emph{real}, \emph{fake audio real visual}, \emph{real audio fake visual}, and \emph{fake audio fake visual}.

\subsection{Perturbation}
\label{ssec:perturb}

Deepfake videos distributing on the Internet are commonly compressed, re‑encoded, streamed through unstable networks, uploaded again after social‑media editing and finally watched on various devices. To close this realism gap \datasetabbr{} includes a wide range of perturbations after the forgery has been applied. The used perturbation methods are provided in \autoref{tab:perturbations}.

\begin{table*}[t]
\centering
\caption{Synthetic perturbations applied to the dataset. Four blank rows are kept at the top for future edits.}
\label{tab:perturbations}
\scalebox{0.83}{\begin{tabular}{lccl}
\toprule[0.4mm]
\rowcolor{mygray}\textbf{Method name} & \textbf{Type} & \textbf{Modality} & \textbf{Explanation}\\
\midrule
 \texttt{VITS}& TTS & Audio & End-to-end speech synthesis combining variational autoencoder, flows, and adversarial training \\
 \texttt{F5TTS}& TTS & Audio & Lightweight, optimized TTS model aimed at real-time speech  \\
 \texttt{XTTSv2}& TTS & Audio & Open-source model for multilingual, cross‑lingual voice cloning \\
 \texttt{YourTTS}& TTS & Audio & Multilingual, multi-speaker TTS enabling zero-shot voice cloning\\
 \midrule
  \texttt{LatentSync}& LipSync & Visual & Audio-conditioned latent diffusion model trained with SyncNet \\
  \texttt{Diff2Lip}& LipSync & Visual & Audio‑conditioned diffusion model that inpaints only mouth region  \\
  \texttt{TalkLip}& LipSync & Visual & Lightweight, real-time talking-face generation model  \\
\midrule
\rowcolor{mygray}\multicolumn{4}{c}{\textbf{Training / Validation}}\\
\midrule
\texttt{GAUSSIAN\_BLUR}          & Perturbation & Visual & Gaussian smoothing that mimics out-of-focus capture.\\
\texttt{SALT\_AND\_PEPPER}       & Perturbation & Visual & Random white/black pixels simulating sensor dust or errors.\\
\texttt{LOW\_BITRATE}           & Perturbation & Visual & Down-/up-scale to create blocky codec artefacts.\\
\texttt{GAUSSIAN\_NOISE}         & Perturbation & Visual & Add zero-mean Gaussian noise typical of sensors.\\
\texttt{POISSON\_NOISE}          & Perturbation & Visual & Photon-count noise via Poisson distribution.\\
\texttt{SPECKLE\_NOISE}          & Perturbation & Visual & Multiplicative granular noise like coherent imaging.\\
\texttt{COLOR\_QUANTIZATION}     & Perturbation & Visual & Reduce palette, producing banding effects.\\
\texttt{RANDOM\_BRIGHTNESS}      & Perturbation & Visual & Random gain/offset to imitate exposure changes.\\
\texttt{MOTION\_BLUR}            & Perturbation & Visual & Linear blur along a direction from camera/object motion.\\
\texttt{ROLLING\_SHUTTER}        & Perturbation & Visual & Row-wise temporal shift causing geometry distortion.\\
\texttt{CAMERA\_SHAKE}           & Perturbation & Visual & Small translational jitters of handheld capture.\\
\texttt{LENS\_DISTORTION}        & Perturbation & Visual & Barrel/pincushion warping from lens aberrations.\\
\texttt{VIGNETTING}              & Perturbation & Visual & Darken edges relative to center (lens fall-off).\\
\texttt{EXPOSURE\_VARIATION}     & Perturbation & Visual & Global gain shift for over/under-exposure.\\
\texttt{CHROMATIC\_ABERRATION}   & Perturbation & Visual & Shift color channels to create fringes.\\
\midrule
\texttt{COMPRESSION\_ARTIFACTS}      & Perturbation & Audio & Quantization noise and high-frequency loss from lossy codecs.\\
\texttt{PITCH\_LOUDNESS}             & Perturbation & Audio & Gain/EQ change emulating device response.\\
\texttt{WHITE\_NOISE}                & Perturbation & Audio & Broadband electronic hiss added to signal.\\
\texttt{TIME\_STRETCH}               & Perturbation & Audio & Change speed without pitch shift (rate variation).\\
\texttt{REVERBERATION}               & Perturbation & Audio & Convolve with room impulse for echoes.\\
\texttt{AMBIENT\_NOISE}              & Perturbation & Audio & Mix environmental sounds (crowd, traffic).\\
\texttt{CLIPPING}                    & Perturbation & Audio & Hard-limit amplitude causing distortion.\\
\texttt{FREQUENCY\_FILTER}           & Perturbation & Audio & Low/High/Band-pass to mimic channel limits.\\
\texttt{DOPPLER}                     & Perturbation & Audio & Time-varying frequency shift from motion.\\
\texttt{INTERFERENCE}                & Perturbation & Audio & Short static bursts emulating electromagnetic noise.\\
\texttt{ROOM\_IMPULSE}               & Perturbation & Audio & Add complex room impulse response for acoustics.\\
\texttt{PAD\_SIMULATION}             & Perturbation & Audio & Simulate padding at the beginning / ending of the clip.\\
\midrule
\rowcolor{mygray}\multicolumn{4}{c}{\textbf{TestA / TestB}}\\
\midrule
\texttt{FRAME\_RATE\_JITTER}         & Perturbation & Visual & Segment-wise FPS variation causing jerky motion.\\
\texttt{PIXELATION\_DISTORTION}      & Perturbation & Visual & Severe local pixelation akin to privacy masks.\\
\texttt{LOCALIZED\_DEFOCUS\_BLUR}    & Perturbation & Visual & Blur only in random spatial regions.\\
\texttt{FRAME\_DROPOUTS}             & Perturbation & Visual & Remove frames, producing temporal jumps.\\
\texttt{RANDOM\_SPATIAL\_WARPING}    & Perturbation & Visual & Subtle random geometric warp per frame.\\
\texttt{RANDOM\_FRAME\_SHUFFLE}      & Perturbation & Visual & Randomly permute contiguous frame chunks, producing temporal disorder.\\
\midrule
\texttt{AUDIO\_STUTTER\_REPEAT}      & Perturbation & Audio & Repeat previous audio frame, producing stutter effect.\\
\texttt{AUDIO\_STUTTER}              & Perturbation & Audio & Repeat short audio segments (buffering).\\
\texttt{AUDIO\_FRAME\_SHUFFLE}       & Perturbation & Audio & Shuffle small audio frame segments to disorder sequence.\\
\texttt{PAD\_SIMULATION}             & Perturbation & Audio & Simulate padding at the beginning / ending of the clip.\\
\bottomrule[0.4mm]
\end{tabular}}
\end{table*}

\subsection{Dataset Splitting}
\label{ssec:split}

For easier reproducing the research in the community, we follow the previous works~\cite{caiYou2022, caiAVDeepfake1M2024, heForgeryNet2021} to pre-defined the dataset splits. For effectively evaluate the performance of the deepfake detection and temporal localization methods, we use two different strategy to split the subsets. Firstly, we split the dataset into three sets: training-validation-combined, testA and testB, with different identities, real sources and the generative methods, to ensure the different domain to evaluate the methods' cross-domain generalizability. For training and validation split, it is randomly split in the sample level, to evaluate the dataset inner domain.

\begin{table*}[t]
\centering
\caption{\textbf{Number of subjects and videos in \datasetabbr{}.} \textmd{``\#" means ``the number of". We show the number of video samples, real samples, fake samples, the number of frames, the total video length and the number of subjects in the table. Note the subjects are shared between \emph{training} and \emph{validation} sets, and \emph{testA} and \emph{testB} sets.}}
\scalebox{1}{
\begin{tabular}{l|cccccc}
\toprule[0.4mm]
\rowcolor{mygray} \textbf{Subset} & \textbf{\#Videos} & \textbf{\#Real} & \textbf{\#Fake} & \textbf{\#Frames} & \textbf{Time (Hour)} & \textbf{\#Subjects} \\ \hline \hline
Training & 1,099,217 & 297,389 & 801,828 & 264,053,153 & 2,444.9 & \multirow{2}{*}{2,606} \\
Validation & 77,326 & 20,220 & 57,106 & 18,488,518 & 171.2 & \\ \hline
TestA & 828,318 & 287,517 & 540,801 & 208,290,429 & 1,928.6 & \multirow{2}{*}{4,503} \\
TestB & 46,293 & 22,810 & 23,483 & 12,012,810 & 111.2 & \\ \hline
Overall & 2,051,154 & 627,936 & 1,423,218 & 502,844,910 & 4,655.9 & 7,109 \\
\bottomrule[0.4mm]
\end{tabular}}
\label{tab:dataset_stats}
\end{table*}

\section{Dataset Statistics \& Analysis}
\label{sec:stats}

\subsection{Scale}
As mentioned in \autoref{ssec:split}, \datasetabbr{} is split into \textit{training}, \textit{validation}, \textit{testA} and \textit{testB} subsets. The detailed statistics about the number of videos, real samples, fake samples, the number of frames, the video length and the number of subjects are displayed in \autoref{tab:dataset_stats}.

Comparing to the previous AV-Deepfake1M only containing 1.1M videos and 2K subjects, \datasetabbr{} significantly exceed the scale (2.1M videos with 7K subjects) and provide more extensive dataset for the community.

\subsection{Deepfake Generation Methods}
One advantage of \datasetabbr{} comparing to previous AV-Deepfake1M~\cite{caiAVDeepfake1M2024} and LAV-DF~\cite{caiYou2022} datasets is using more generation methods. We calculate the statistics and they are shown in \autoref{fig:dataset_comparison}.

LAV-DF uses only one method Wav2Lip~\cite{prajwalLip2020} for generating visual frames and single method SV2TTS~\cite{jiaTransfer2018} for generating fake audio, which is limited to the diversity of the low-level artifacts to be detected. AV-Deepfake1M dramatically improve the generation quality and the fake samples are difficult to be noticed by human based on their user study. However, due to the limited generation methods, the trained deep learning models can catch the low level, and become more effective than the human performance~\cite{zhangMFMS2024, perez-vieitesVigo2024, wangBuilding2024}. \datasetabbr{} overcome the generation methods diversity issue by involving more generative methods (LatentSync~\cite{liLatentSync2025}, Diff2Lip~\cite{mukhopadhyayDiff2Lip2024}, F5TTS~\cite{chenF5TTS2025}, XTTSv2~\cite{casanovaXTTS2024}).

\subsection{Perturbation Methods}

\begin{figure*}[t]
\centering
\includegraphics[width=\linewidth]{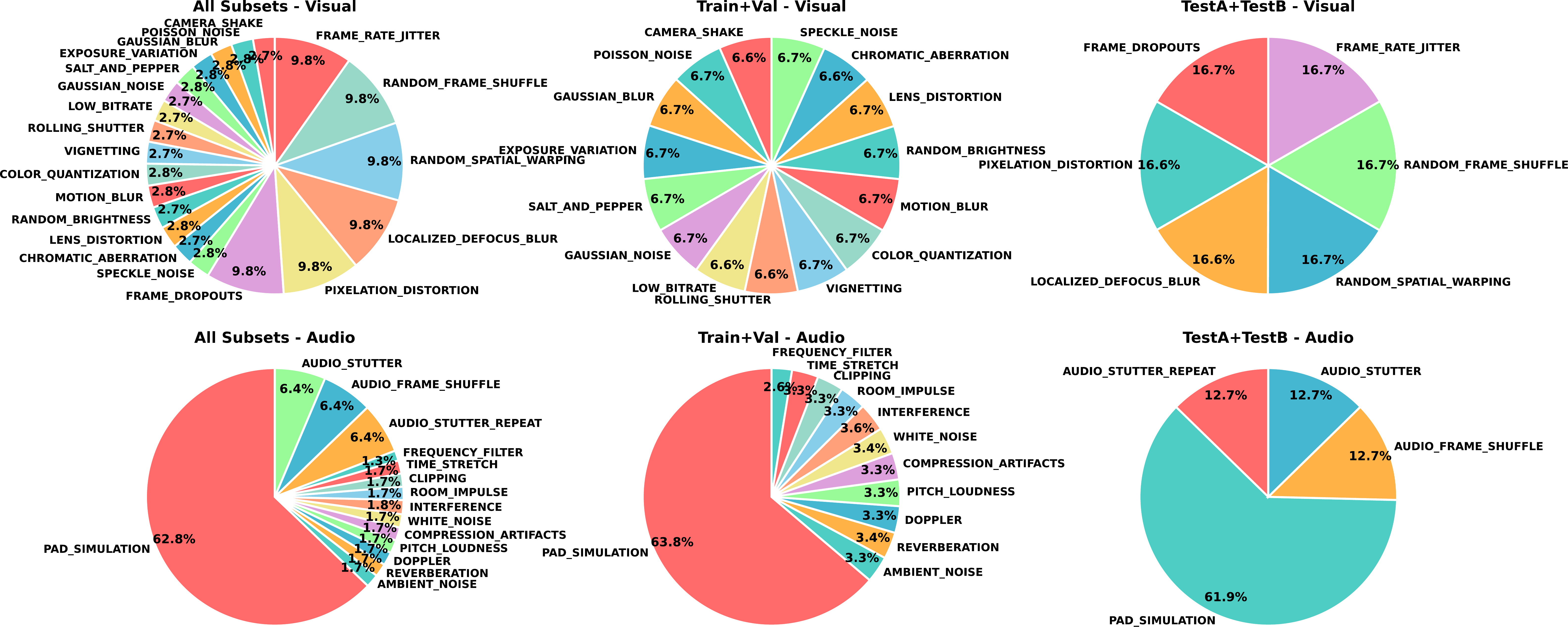}
\caption{\textbf{The distribution of audio and visual perturbations in \datasetabbr{}.} \textmd{The first row of pie charts shows the perturbations in the visual modality. The second row shows the audio modality. The first column shows the perturbations in the whole dataset, and the second, third columns show the perturbations in the different subsets.}}
\label{fig:dataset_purturbation}
\end{figure*}

In \autoref{fig:dataset_purturbation}, we show the distribution of perturbations we used in each modality and subsets. Each video can contain zero, one, or multiple perturbations. For \emph{training} and \emph{validation} subsets, we use a different set of perturbations compared to \emph{testA} and \emph{testB} subsets. By isolating the perturbations in the different subsets, the methods to be evaluated should be capable to understand the concept of fake manipulations and real perturbations. 

\begin{figure}[h]
\centering
\includegraphics[width=\linewidth]{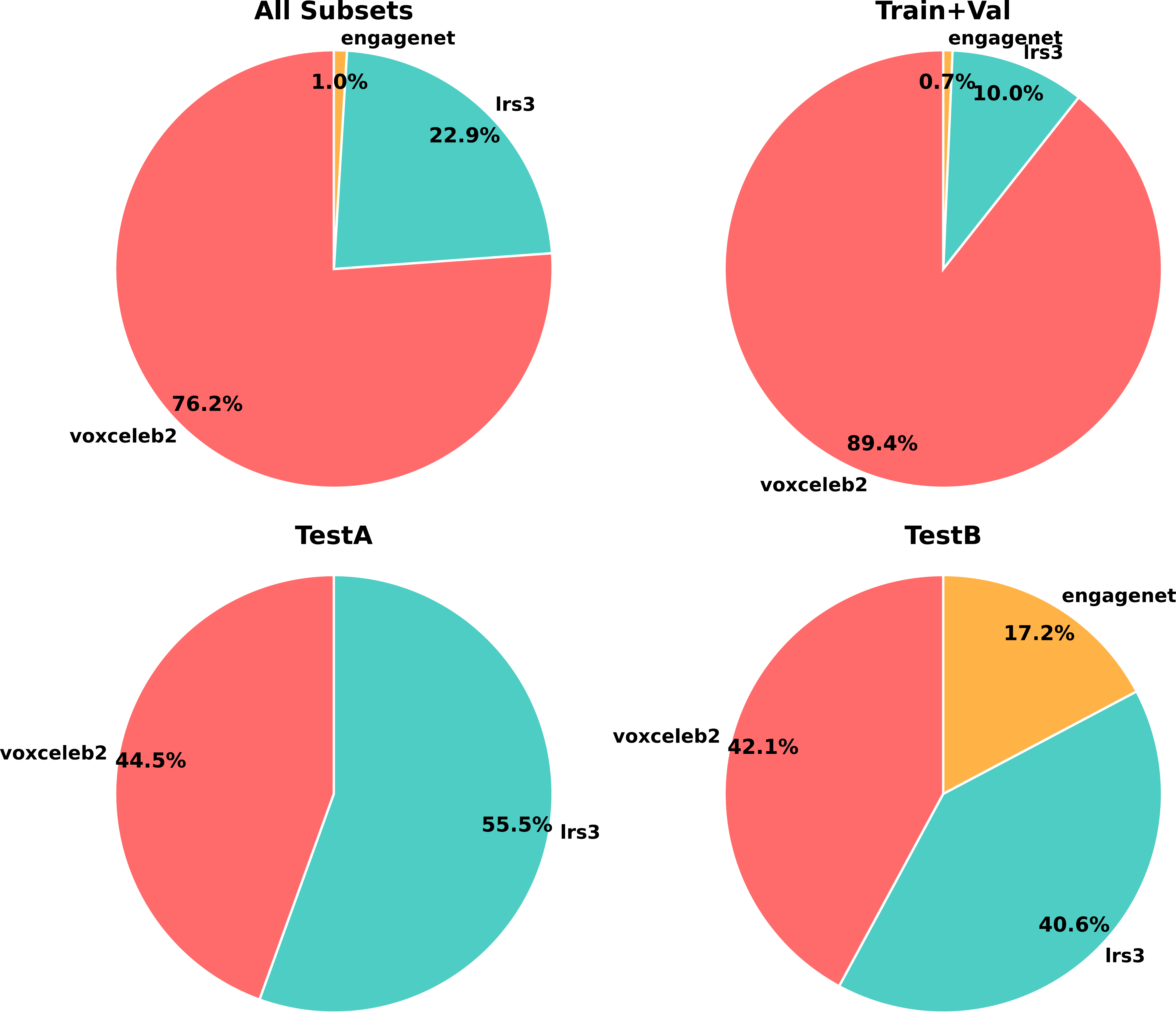}
\caption{\textbf{The proportion statistics of source dataset in \datasetabbr{}.} \textmd{We show the the source dataset for the whole dataset, and different subsets.}}
\label{fig:dataset_source}
\end{figure}

\subsection{Real Video Sources}

\datasetabbr{} is generated based on the multiple real dataset sources, including VoxCeleb2~\cite{chungVoxCeleb22018}, LRS3~\cite{afourasLRS3TED2018} and EngageNet~\cite{singhHave2023}. The proportion of each data source in different subsets is shown in \autoref{fig:dataset_source}. Comparing to previous datasets~\cite{caiYou2022, caiAVDeepfake1M2024} only using the VoxCeleb2 as the data source, the extra data source in \datasetabbr{} provides more diversity to the dataset for training and benchmarking the methods.

% -------------------------------------------------------------------------
\section{Challenge}
\label{sec:exp}

Based on the proposed method, we host 2025 1M-Deepfakes Detection Challenge at ACM Multimedia conference. In this section, we report the benchmark of several baseline methods~\cite{caiYou2022, caiGlitch2023, cholletXception2017} and top teams. Please refer to the challenge leaderboard page for more details\footnote{\url{https://deepfakes1m.github.io/2025/evaluation}}. All three baselines were trained on \datasetabbr{}. The source code for implementing these baselines is available in our GitHub repository\footnote{\url{https://github.com/ControlNet/AV-Deepfake1M}}.

\subsection{Benchmark Protocol}
\label{ssec:protocol}

All methods are trained only on the official \emph{training} split (\autoref{tab:dataset_stats}) and evaluated on \emph{TestA} and \emph{TestB} subsets. We follow the same evaluation metrics as the challenge in the last year~\cite{cai1MDeepfakes2024}, the same metrics are used 
\emph{AUC} for classification; an averaged score of \emph{AP}@\{0.50,0.75,0.90,0.95\}
and \textit{AR}@\{50,30,20,10,5\} for localization.

%------------------------------------------------------------------------
\subsection{Quantitative Results}
\label{ssec:quantitative}

\paragraph{Video‑level classification.}
\autoref{tab:qual_cls} shows the leaderboard of the challenge. The best team (XJTU SunFlower Lab) produces
an impressive \textbf{0.9783}~AUC, yet the Xception baseline reaches only \textbf{0.5509}.

\paragraph{Temporal localization.}
\autoref{tab:qual_loc} shows that the top team
\emph{Pindrop Labs} surpasses BA‑TFD+ by
\(\approx\) \textbf{0.52} of the localization score.
Even BA‑TFD+~\cite{caiGlitch2023}, which scored 96.30 AP@0.5 on previous LAV-DF dataset~\cite{caiYou2022}, now struggles at 14.7 (AP@0.5).  Such a dramatic collapse highlights how the new perturbations and synthesis pipelines invalidate the method design that performs well on AV‑Deepfake1M~\cite{caiAVDeepfake1M2024} and LAV‑DF~\cite{caiYou2022}. However, not like the classification results are closed to saturated, there is potential for the community to push the performance for temporal localization task.

\begin{table}[t]
\centering
\caption{\textbf{Quantitative result of the \textit{TestA}
classification (AUC).}}
\label{tab:qual_cls}
\scalebox{1}{
\begin{tabular}{lcc}
\toprule[0.4mm]
\rowcolor{mygray}\textbf{Team / Method} & \textbf{TestA} & \textbf{TestB}\\
\midrule
XJTU SunFlower Lab & 97.83 & -\\
WHU\_SPEECH        & 93.07 & -\\
KLASS              & 92.78 & -\\
Pindrop Labs       & 92.49 & -\\
Mizhi Labs         & 91.78 & -\\\hline
Xception (baseline)~\cite{cholletXception2017} & 55.09 & 57.29\\
\bottomrule[0.4mm]
\end{tabular}}
\end{table}

\begin{table*}[t]
\centering
\caption{\textbf{Quantitative result of temporal localization task on \emph{testA} subset}.}
\label{tab:qual_loc}
\scalebox{1}{
\begin{tabular}{lcccccccccc}
\toprule[0.4mm]
\rowcolor{mygray}\textbf{Team / Method} & \textbf{Score} & \textbf{AP@0.5} & \textbf{AP@0.75} & \textbf{AP@0.9} & \textbf{AP@0.95} & \textbf{AR@50} & \textbf{AR@30} & \textbf{AR@20} & \textbf{AR@10} & \textbf{AR@5}\\
\midrule
Pindrop Labs            & 67.20 & 77.94 & 66.52 & 44.66 & 34.27 & 80.30 & 80.09 & 79.59 & 77.84 & 74.93\\
Mizhi Lab               & 55.00 & 72.81 & 58.30 & 32.68 & 15.46 & 65.20 & 65.20 & 65.20 & 65.20 & 65.14\\
Purdue‑M2               & 50.87 & 62.62 & 52.49 & 43.76 & 26.16 & 55.56 & 55.56 & 55.56 & 55.52 & 55.20\\
WHU\_SPEECH             & 41.30 & 50.52 & 34.38 & 12.58 &  04.25 & 60.27 & 58.87 & 57.50 & 55.51 & 53.64\\
KLASS                   & 35.36 & 51.17 & 40.17 & 17.01 &  04.16 & 42.59 & 42.59 & 42.59 & 42.59 & 42.58\\\hline
BA‑TFD+ (baseline)~\cite{caiGlitch2023} & 14.71 & 14.01 &  02.35 & 00.05 & 00.00 & 32.80 & 30.17 & 26.61 & 20.88 & 16.11\\
BA‑TFD (baseline)~\cite{caiYou2022}     & 13.54 & 09.81 & 01.29 & 00.04 & 00.00 & 33.25 & 29.16 & 25.20 & 19.24 & 14.59\\
\bottomrule[0.4mm]
\end{tabular}}
\end{table*}

\begin{table*}[t]
\centering
\caption{\textbf{Quantitative result of temporal localization task on \emph{testB} subset}.}
\label{tab:qual_loc_testB}
\scalebox{1}{
\begin{tabular}{lcccccccccc}
\toprule[0.4mm]
\rowcolor{mygray}\textbf{Team / Method} & \textbf{Score} & \textbf{AP@0.5} & \textbf{AP@0.75} & \textbf{AP@0.9} & \textbf{AP@0.95} & \textbf{AR@50} & \textbf{AR@30} & \textbf{AR@20} & \textbf{AR@10} & \textbf{AR@5}\\
\midrule
% Pindrop Labs            & \\
% Mizhi Lab               & \\
% Purdue‑M2               & \\
% WHU\_SPEECH             & \\
% KLASS                   & \\\hline
BA‑TFD+ (baseline)~\cite{caiGlitch2023} & 15.15 & 16.20 & 03.39 & 00.12 & 00.01 & 32.45 & 29.35 & 26.51 & 21.56 & 16.98 \\
BA‑TFD (baseline)~\cite{caiYou2022}     & 11.17 & 04.63 & 00.57 & 00.02 & 00.00 & 28.71 & 25.14 & 22.03 & 16.81 & 12.50\\
\bottomrule[0.4mm]
\end{tabular}}
\end{table*}

%------------------------------------------------------------------------
% \subsection{Take‑away messages}

% \noindent\textbf{(1)} \datasetabbr{} breaks the implicit
% assumptions of current detectors; \textbf{(2)} new research must
% explicitly model realistic perturbations and multi‑modal temporal
% consistency; \textbf{(3)} even simple baselines provide valuable
% sandbox reference points for the community.

% -------------------------------------------------------------------------
\section{Conclusion}
\label{sec:conclusion}

We have presented \textbf{\datasetabbr{}}, a new large–scale benchmark contributing to the audio-visual deepfake research in three key dimensions: \textit{scale}, \textit{generation diversity}, and \textit{real-world perturbations}. Together with a evaluation protocol and baselines, the benchmark underpinned the \emph{1M-Deepfakes Detection Challenge 2025}, whose results reveal substantial performance gaps, especially for temporal localization once detectors are confronted with unseen synthesis methods and distribution artifacts. 

\noindent\textbf{Future directions.} Based on the experience in creating \datasetabbr{} and its experiments, we see following important future directions:

\begin{itemize}[leftmargin=1.2em,itemsep=2pt]
    \item \textbf{Deployment and Explainability.} For large-scale deployment of deepfake detectors, it is important to explain why the system classifies a given input as manipulated. Additionally, effective strategies are needed to ensure these explanations are understandable to non-technical users. An important question is: \textbf{How can deepfake detection and its associated explanations be made more accessible and user-friendly for a broad audience?} A recent approach toward generating simpler explanations using text and images is proposed in \cite{narang2025laylens}.
    
    \item \textbf{Perturbation-robust representation learning.} New training objectives and augmentation strategies are needed to disentangle semantic manipulation from perturbations such as compression, noise, or frame-rate jitter.
    
    \item \textbf{Rapid adaptation to novel forgery pipelines.}  Few-shot and continual-learning techniques could enable detectors to track the fast-moving frontier of diffusion- and LLM-driven generators without exhaustive re-training.
    
    \item \textbf{Fine-grained multimodal reasoning.}  Beyond low-level artifacts in the video, future methods should jointly understand the high-level context of fake videos for reasoning.
    
    \item \textbf{Cross-cultural and multilingual robustness.}  As manipulation semantics vary with language and culture, detectors and benchmarks must cover a broader linguistic landscape and account for culturally specific rhetorical cues \cite{kuckrejaTell2025}.
    
    \item \textbf{Open-world evaluation.} The future deepfake detectors should be robust and generalizable for unseen generation methods and perturbations (\ie open-set conditions).
    
    \item \textbf{Ethics, fairness and privacy.} Large-scale dataset collection and release of manipulated media has potential risk of privacy leakage and misuse. This concerns can be addressed by the simulated and synthetic data generation, and migrated the detector for the real use.
\end{itemize}

We hope that \datasetabbr{}, with its breadth of sources, manipulations and perturbations, will become a cornerstone benchmark, fostering robust, generalizable, and socially responsible solutions to the ever-evolving deepfake threat.

\clearpage

\bibliographystyle{ACM-Reference-Format}
\bibliography{Bibliography}


\begin{thebibliography}{45}

%%% the Web address.

\ifx \showCODEN    \undefined \def \showCODEN     #1{\unskip}     \fi
\ifx \showISBNx    \undefined \def \showISBNx     #1{\unskip}     \fi
\ifx \showISBNxiii \undefined \def \showISBNxiii  #1{\unskip}     \fi
\ifx \showISSN     \undefined \def \showISSN      #1{\unskip}     \fi
\ifx \showLCCN     \undefined \def \showLCCN      #1{\unskip}     \fi
\ifx \shownote     \undefined \def \shownote      #1{#1}          \fi
\ifx \showarticletitle \undefined \def \showarticletitle #1{#1}   \fi
\ifx \showURL      \undefined \def \showURL       {\relax}        \fi
% invisible to TeX
\providecommand\bibfield[2]{#2}
\providecommand\bibinfo[2]{#2}
\providecommand\natexlab[1]{#1}
\providecommand\showeprint[2][]{arXiv:#2}

\bibitem[Afouras et~al\mbox{.}(2018)]%
        {afourasLRS3TED2018}
\bibfield{author}{\bibinfo{person}{Triantafyllos Afouras}, \bibinfo{person}{Joon~Son Chung}, {and} \bibinfo{person}{Andrew Zisserman}.} \bibinfo{year}{2018}\natexlab{}.
\newblock \bibinfo{title}{{LRS3}-{TED}: a large-scale dataset for visual speech recognition}.
\newblock
\href{https://doi.org/10.48550/arXiv.1809.00496}{doi:\nolinkurl{10.48550/arXiv.1809.00496}}
\newblock
\shownote{arXiv:1809.00496 [cs]}.


\bibitem[Cai et~al\mbox{.}(2024a)]%
        {cai1MDeepfakes2024}
\bibfield{author}{\bibinfo{person}{Zhixi Cai}, \bibinfo{person}{Abhinav Dhall}, \bibinfo{person}{Shreya Ghosh}, \bibinfo{person}{Munawar Hayat}, \bibinfo{person}{Dimitrios Kollias}, \bibinfo{person}{Kalin Stefanov}, {and} \bibinfo{person}{Usman Tariq}.} \bibinfo{year}{2024}\natexlab{a}.
\newblock \showarticletitle{{1M}-{Deepfakes} {Detection} {Challenge}}. In \bibinfo{booktitle}{\emph{Proceedings of the 32nd {ACM} {International} {Conference} on {Multimedia}}} \emph{(\bibinfo{series}{{MM} '24})}. \bibinfo{publisher}{Association for Computing Machinery}, \bibinfo{address}{New York, NY, USA}, \bibinfo{pages}{11355--11359}.
\newblock
\showISBNx{9798400706868}
\href{https://doi.org/10.1145/3664647.3689145}{doi:\nolinkurl{10.1145/3664647.3689145}}


\bibitem[Cai et~al\mbox{.}(2024b)]%
        {caiAVDeepfake1M2024}
\bibfield{author}{\bibinfo{person}{Zhixi Cai}, \bibinfo{person}{Shreya Ghosh}, \bibinfo{person}{Aman~Pankaj Adatia}, \bibinfo{person}{Munawar Hayat}, \bibinfo{person}{Abhinav Dhall}, \bibinfo{person}{Tom Gedeon}, {and} \bibinfo{person}{Kalin Stefanov}.} \bibinfo{year}{2024}\natexlab{b}.
\newblock \showarticletitle{{AV}-{Deepfake1M}: {A} {Large}-{Scale} {LLM}-{Driven} {Audio}-{Visual} {Deepfake} {Dataset}}. In \bibinfo{booktitle}{\emph{Proceedings of the 32nd {ACM} {International} {Conference} on {Multimedia}}} \emph{(\bibinfo{series}{{MM} '24})}. \bibinfo{publisher}{Association for Computing Machinery}, \bibinfo{address}{New York, NY, USA}, \bibinfo{pages}{7414--7423}.
\newblock
\showISBNx{9798400706868}
\href{https://doi.org/10.1145/3664647.3680795}{doi:\nolinkurl{10.1145/3664647.3680795}}


\bibitem[Cai et~al\mbox{.}(2023)]%
        {caiGlitch2023}
\bibfield{author}{\bibinfo{person}{Zhixi Cai}, \bibinfo{person}{Shreya Ghosh}, \bibinfo{person}{Abhinav Dhall}, \bibinfo{person}{Tom Gedeon}, \bibinfo{person}{Kalin Stefanov}, {and} \bibinfo{person}{Munawar Hayat}.} \bibinfo{year}{2023}\natexlab{}.
\newblock \showarticletitle{Glitch in the matrix: {A} large scale benchmark for content driven audio–visual forgery detection and localization}.
\newblock \bibinfo{journal}{\emph{Computer Vision and Image Understanding}}  \bibinfo{volume}{236} (\bibinfo{date}{Nov.} \bibinfo{year}{2023}), \bibinfo{pages}{103818}.
\newblock
\showISSN{1077-3142}
\href{https://doi.org/10.1016/j.cviu.2023.103818}{doi:\nolinkurl{10.1016/j.cviu.2023.103818}}


\bibitem[Cai et~al\mbox{.}(2022)]%
        {caiYou2022}
\bibfield{author}{\bibinfo{person}{Zhixi Cai}, \bibinfo{person}{Kalin Stefanov}, \bibinfo{person}{Abhinav Dhall}, {and} \bibinfo{person}{Munawar Hayat}.} \bibinfo{year}{2022}\natexlab{}.
\newblock \showarticletitle{Do {You} {Really} {Mean} {That}? {Content} {Driven} {Audio}-{Visual} {Deepfake} {Dataset} and {Multimodal} {Method} for {Temporal} {Forgery} {Localization}}. In \bibinfo{booktitle}{\emph{2022 {International} {Conference} on {Digital} {Image} {Computing}: {Techniques} and {Applications} ({DICTA})}}. \bibinfo{address}{Sydney, Australia}, \bibinfo{pages}{1--10}.
\newblock
\href{https://doi.org/10.1109/DICTA56598.2022.10034605}{doi:\nolinkurl{10.1109/DICTA56598.2022.10034605}}


\bibitem[Casanova et~al\mbox{.}(2024)]%
        {casanovaXTTS2024}
\bibfield{author}{\bibinfo{person}{Edresson Casanova}, \bibinfo{person}{Kelly Davis}, \bibinfo{person}{Eren Gölge}, \bibinfo{person}{Görkem Göknar}, \bibinfo{person}{Iulian Gulea}, \bibinfo{person}{Logan Hart}, \bibinfo{person}{Aya Aljafari}, \bibinfo{person}{Joshua Meyer}, \bibinfo{person}{Reuben Morais}, \bibinfo{person}{Samuel Olayemi}, {and} \bibinfo{person}{Julian Weber}.} \bibinfo{year}{2024}\natexlab{}.
\newblock \bibinfo{title}{{XTTS}: a {Massively} {Multilingual} {Zero}-{Shot} {Text}-to-{Speech} {Model}}.
\newblock
\href{https://doi.org/10.48550/arXiv.2406.04904}{doi:\nolinkurl{10.48550/arXiv.2406.04904}}
\newblock
\shownote{arXiv:2406.04904 [cs, eess]}.


\bibitem[Casanova et~al\mbox{.}(2022)]%
        {casanovaYourTTS2022}
\bibfield{author}{\bibinfo{person}{Edresson Casanova}, \bibinfo{person}{Julian Weber}, \bibinfo{person}{Christopher~D. Shulby}, \bibinfo{person}{Arnaldo~Candido Junior}, \bibinfo{person}{Eren Gölge}, {and} \bibinfo{person}{Moacir~A. Ponti}.} \bibinfo{year}{2022}\natexlab{}.
\newblock \showarticletitle{{YourTTS}: {Towards} {Zero}-{Shot} {Multi}-{Speaker} {TTS} and {Zero}-{Shot} {Voice} {Conversion} for {Everyone}}. In \bibinfo{booktitle}{\emph{Proceedings of the 39th {International} {Conference} on {Machine} {Learning}}}. \bibinfo{publisher}{PMLR}, \bibinfo{pages}{2709--2720}.
\newblock
\urldef\tempurl%
\url{https://proceedings.mlr.press/v162/casanova22a.html}
\showURL{%
\tempurl}
\newblock
\shownote{ISSN: 2640-3498}.


\bibitem[Chen et~al\mbox{.}(2025)]%
        {chenF5TTS2025}
\bibfield{author}{\bibinfo{person}{Yushen Chen}, \bibinfo{person}{Zhikang Niu}, \bibinfo{person}{Ziyang Ma}, \bibinfo{person}{Keqi Deng}, \bibinfo{person}{Chunhui Wang}, \bibinfo{person}{Jian Zhao}, \bibinfo{person}{Kai Yu}, {and} \bibinfo{person}{Xie Chen}.} \bibinfo{year}{2025}\natexlab{}.
\newblock \bibinfo{title}{F5-{TTS}: {A} {Fairytaler} that {Fakes} {Fluent} and {Faithful} {Speech} with {Flow} {Matching}}.
\newblock
\href{https://doi.org/10.48550/arXiv.2410.06885}{doi:\nolinkurl{10.48550/arXiv.2410.06885}}
\newblock
\shownote{arXiv:2410.06885 [eess]}.


\bibitem[Chollet(2017)]%
        {cholletXception2017}
\bibfield{author}{\bibinfo{person}{Francois Chollet}.} \bibinfo{year}{2017}\natexlab{}.
\newblock \showarticletitle{Xception: {Deep} {Learning} {With} {Depthwise} {Separable} {Convolutions}}. In \bibinfo{booktitle}{\emph{Proceedings of the {IEEE} {Conference} on {Computer} {Vision} and {Pattern} {Recognition}}}. \bibinfo{pages}{1251--1258}.
\newblock
\urldef\tempurl%
\url{https://openaccess.thecvf.com/content_cvpr_2017/html/Chollet_Xception_Deep_Learning_CVPR_2017_paper.html}
\showURL{%
\tempurl}


\bibitem[Chung et~al\mbox{.}(2018)]%
        {chungVoxCeleb22018}
\bibfield{author}{\bibinfo{person}{Joon~Son Chung}, \bibinfo{person}{Arsha Nagrani}, {and} \bibinfo{person}{Andrew Zisserman}.} \bibinfo{year}{2018}\natexlab{}.
\newblock \showarticletitle{{VoxCeleb2}: {Deep} {Speaker} {Recognition}}. In \bibinfo{booktitle}{\emph{Interspeech 2018}}. \bibinfo{publisher}{ISCA}, \bibinfo{pages}{1086--1090}.
\newblock
\href{https://doi.org/10.21437/Interspeech.2018-1929}{doi:\nolinkurl{10.21437/Interspeech.2018-1929}}


\bibitem[Dolhansky et~al\mbox{.}(2020)]%
        {dolhanskyDeepFake2020}
\bibfield{author}{\bibinfo{person}{Brian Dolhansky}, \bibinfo{person}{Joanna Bitton}, \bibinfo{person}{Ben Pflaum}, \bibinfo{person}{Jikuo Lu}, \bibinfo{person}{Russ Howes}, \bibinfo{person}{Menglin Wang}, {and} \bibinfo{person}{Cristian~Canton Ferrer}.} \bibinfo{year}{2020}\natexlab{}.
\newblock \bibinfo{title}{The {DeepFake} {Detection} {Challenge} ({DFDC}) {Dataset}}.
\newblock
\urldef\tempurl%
\url{http://arxiv.org/abs/2006.07397}
\showURL{%
\tempurl}
\newblock
\shownote{arXiv: 2006.07397 [cs]}.


\bibitem[Défossez et~al\mbox{.}(2020)]%
        {defossezReal2020}
\bibfield{author}{\bibinfo{person}{Alexandre Défossez}, \bibinfo{person}{Gabriel Synnaeve}, {and} \bibinfo{person}{Yossi Adi}.} \bibinfo{year}{2020}\natexlab{}.
\newblock \showarticletitle{Real {Time} {Speech} {Enhancement} in the {Waveform} {Domain}}. In \bibinfo{booktitle}{\emph{Interspeech 2020}}. \bibinfo{address}{Shanghai, China}, \bibinfo{pages}{3291--3295}.
\newblock
\href{https://doi.org/10.21437/Interspeech.2020-2409}{doi:\nolinkurl{10.21437/Interspeech.2020-2409}}


\bibitem[Gupta et~al\mbox{.}(2025)]%
        {guptaMultiverse2025}
\bibfield{author}{\bibinfo{person}{Parul Gupta}, \bibinfo{person}{Shreya Ghosh}, \bibinfo{person}{Tom Gedeon}, \bibinfo{person}{Thanh-Toan Do}, {and} \bibinfo{person}{Abhinav Dhall}.} \bibinfo{year}{2025}\natexlab{}.
\newblock \bibinfo{title}{Multiverse {Through} {Deepfakes}: {The} {MultiFakeVerse} {Dataset} of {Person}-{Centric} {Visual} and {Conceptual} {Manipulations}}.
\newblock
\href{https://doi.org/10.48550/arXiv.2506.00868}{doi:\nolinkurl{10.48550/arXiv.2506.00868}}
\newblock
\shownote{arXiv:2506.00868 [cs]}.


\bibitem[He et~al\mbox{.}(2021)]%
        {heForgeryNet2021}
\bibfield{author}{\bibinfo{person}{Yinan He}, \bibinfo{person}{Bei Gan}, \bibinfo{person}{Siyu Chen}, \bibinfo{person}{Yichun Zhou}, \bibinfo{person}{Guojun Yin}, \bibinfo{person}{Luchuan Song}, \bibinfo{person}{Lu Sheng}, \bibinfo{person}{Jing Shao}, {and} \bibinfo{person}{Ziwei Liu}.} \bibinfo{year}{2021}\natexlab{}.
\newblock \showarticletitle{{ForgeryNet}: {A} {Versatile} {Benchmark} for {Comprehensive} {Forgery} {Analysis}}. In \bibinfo{booktitle}{\emph{Proceedings of the {IEEE}/{CVF} {Conference} on {Computer} {Vision} and {Pattern} {Recognition}}}. \bibinfo{pages}{4360--4369}.
\newblock
\urldef\tempurl%
\url{https://openaccess.thecvf.com/content/CVPR2021/html/He_ForgeryNet_A_Versatile_Benchmark_for_Comprehensive_Forgery_Analysis_CVPR_2021_paper.html}
\showURL{%
\tempurl}


\bibitem[Hou et~al\mbox{.}(2025)]%
        {houPolyGlotFake2025}
\bibfield{author}{\bibinfo{person}{Yang Hou}, \bibinfo{person}{Haitao Fu}, \bibinfo{person}{Chunkai Chen}, \bibinfo{person}{Zida Li}, \bibinfo{person}{Haoyu Zhang}, {and} \bibinfo{person}{Jianjun Zhao}.} \bibinfo{year}{2025}\natexlab{}.
\newblock \showarticletitle{{PolyGlotFake}: {A} {Novel} {Multilingual} and {Multimodal} {DeepFake} {Dataset}}. In \bibinfo{booktitle}{\emph{Pattern {Recognition}}}, \bibfield{editor}{\bibinfo{person}{Apostolos Antonacopoulos}, \bibinfo{person}{Subhasis Chaudhuri}, \bibinfo{person}{Rama Chellappa}, \bibinfo{person}{Cheng-Lin Liu}, \bibinfo{person}{Saumik Bhattacharya}, {and} \bibinfo{person}{Umapada Pal}} (Eds.). \bibinfo{publisher}{Springer Nature Switzerland}, \bibinfo{address}{Cham}, \bibinfo{pages}{180--193}.
\newblock
\showISBNx{978-3-031-78341-8}
\href{https://doi.org/10.1007/978-3-031-78341-8_12}{doi:\nolinkurl{10.1007/978-3-031-78341-8_12}}


\bibitem[Huang et~al\mbox{.}(2025)]%
        {huangSIDA2025}
\bibfield{author}{\bibinfo{person}{Zhenglin Huang}, \bibinfo{person}{Jinwei Hu}, \bibinfo{person}{Xiangtai Li}, \bibinfo{person}{Yiwei He}, \bibinfo{person}{Xingyu Zhao}, \bibinfo{person}{Bei Peng}, \bibinfo{person}{Baoyuan Wu}, \bibinfo{person}{Xiaowei Huang}, {and} \bibinfo{person}{Guangliang Cheng}.} \bibinfo{year}{2025}\natexlab{}.
\newblock \showarticletitle{{SIDA}: {Social} {Media} {Image} {Deepfake} {Detection}, {Localization} and {Explanation} with {Large} {Multimodal} {Model}}. In \bibinfo{booktitle}{\emph{Proceedings of the {Computer} {Vision} and {Pattern} {Recognition} {Conference}}}. \bibinfo{pages}{28831--28841}.
\newblock
\urldef\tempurl%
\url{https://openaccess.thecvf.com/content/CVPR2025/html/Huang_SIDA_Social_Media_Image_Deepfake_Detection_Localization_and_Explanation_with_CVPR_2025_paper.html}
\showURL{%
\tempurl}


\bibitem[Jia et~al\mbox{.}(2018)]%
        {jiaTransfer2018}
\bibfield{author}{\bibinfo{person}{Ye Jia}, \bibinfo{person}{Yu Zhang}, \bibinfo{person}{Ron~J. Weiss}, \bibinfo{person}{Quan Wang}, \bibinfo{person}{Jonathan Shen}, \bibinfo{person}{Fei Ren}, \bibinfo{person}{Zhifeng Chen}, \bibinfo{person}{Patrick Nguyen}, \bibinfo{person}{Ruoming Pang}, \bibinfo{person}{Ignacio~Lopez Moreno}, {and} \bibinfo{person}{Yonghui Wu}.} \bibinfo{year}{2018}\natexlab{}.
\newblock \showarticletitle{Transfer learning from speaker verification to multispeaker text-to-speech synthesis}. In \bibinfo{booktitle}{\emph{Proceedings of the 32nd {International} {Conference} on {Neural} {Information} {Processing} {Systems}}} \emph{(\bibinfo{series}{{NIPS}'18})}. \bibinfo{publisher}{Curran Associates Inc.}, \bibinfo{address}{Red Hook, NY, USA}, \bibinfo{pages}{4485--4495}.
\newblock


\bibitem[Jiang et~al\mbox{.}(2020)]%
        {jiangDeeperForensics12020}
\bibfield{author}{\bibinfo{person}{Liming Jiang}, \bibinfo{person}{Ren Li}, \bibinfo{person}{Wayne Wu}, \bibinfo{person}{Chen Qian}, {and} \bibinfo{person}{Chen~Change Loy}.} \bibinfo{year}{2020}\natexlab{}.
\newblock \showarticletitle{{DeeperForensics}-1.0: {A} {Large}-{Scale} {Dataset} for {Real}-{World} {Face} {Forgery} {Detection}}. In \bibinfo{booktitle}{\emph{Proceedings of the {IEEE}/{CVF} {Conference} on {Computer} {Vision} and {Pattern} {Recognition}}}. \bibinfo{pages}{2889--2898}.
\newblock
\urldef\tempurl%
\url{https://openaccess.thecvf.com/content_CVPR_2020/html/Jiang_DeeperForensics-1.0_A_Large-Scale_Dataset_for_Real-World_Face_Forgery_Detection_CVPR_2020_paper.html}
\showURL{%
\tempurl}


\bibitem[Khalid et~al\mbox{.}(2021)]%
        {khalidFakeAVCeleb2021}
\bibfield{author}{\bibinfo{person}{Hasam Khalid}, \bibinfo{person}{Shahroz Tariq}, \bibinfo{person}{Minha Kim}, {and} \bibinfo{person}{Simon~S. Woo}.} \bibinfo{year}{2021}\natexlab{}.
\newblock \showarticletitle{{FakeAVCeleb}: {A} {Novel} {Audio}-{Video} {Multimodal} {Deepfake} {Dataset}}. In \bibinfo{booktitle}{\emph{Thirty-fifth {Conference} on {Neural} {Information} {Processing} {Systems} {Datasets} and {Benchmarks} {Track}}}.
\newblock
\urldef\tempurl%
\url{https://openreview.net/forum?id=TAXFsg6ZaOl}
\showURL{%
\tempurl}


\bibitem[Kim et~al\mbox{.}(2021)]%
        {kimConditional2021}
\bibfield{author}{\bibinfo{person}{Jaehyeon Kim}, \bibinfo{person}{Jungil Kong}, {and} \bibinfo{person}{Juhee Son}.} \bibinfo{year}{2021}\natexlab{}.
\newblock \showarticletitle{Conditional {Variational} {Autoencoder} with {Adversarial} {Learning} for {End}-to-{End} {Text}-to-{Speech}}. In \bibinfo{booktitle}{\emph{Proceedings of the 38th {International} {Conference} on {Machine} {Learning}}}. \bibinfo{publisher}{PMLR}, \bibinfo{pages}{5530--5540}.
\newblock
\urldef\tempurl%
\url{https://proceedings.mlr.press/v139/kim21f.html}
\showURL{%
\tempurl}
\newblock
\shownote{ISSN: 2640-3498}.


\bibitem[Korshunov and Marcel(2018)]%
        {korshunovDeepFakes2018}
\bibfield{author}{\bibinfo{person}{Pavel Korshunov} {and} \bibinfo{person}{Sebastien Marcel}.} \bibinfo{year}{2018}\natexlab{}.
\newblock \bibinfo{title}{{DeepFakes}: a {New} {Threat} to {Face} {Recognition}? {Assessment} and {Detection}}.
\newblock
\urldef\tempurl%
\url{http://arxiv.org/abs/1812.08685}
\showURL{%
\tempurl}
\newblock
\shownote{arXiv:1812.08685 [cs]}.


\bibitem[Kuckreja et~al\mbox{.}(2025)]%
        {kuckrejaTell2025}
\bibfield{author}{\bibinfo{person}{Kartik Kuckreja}, \bibinfo{person}{Parul Gupta}, \bibinfo{person}{Injy Hamed}, \bibinfo{person}{Thamar Solorio}, \bibinfo{person}{Muhammad~Haris Khan}, {and} \bibinfo{person}{Abhinav Dhall}.} \bibinfo{year}{2025}\natexlab{}.
\newblock \bibinfo{title}{Tell me {Habibi}, is it {Real} or {Fake}?}
\newblock
\href{https://doi.org/10.48550/arXiv.2505.22581}{doi:\nolinkurl{10.48550/arXiv.2505.22581}}
\newblock
\shownote{arXiv:2505.22581 [cs]}.


\bibitem[Kwon et~al\mbox{.}(2021)]%
        {kwonKoDF2021}
\bibfield{author}{\bibinfo{person}{Patrick Kwon}, \bibinfo{person}{Jaeseong You}, \bibinfo{person}{Gyuhyeon Nam}, \bibinfo{person}{Sungwoo Park}, {and} \bibinfo{person}{Gyeongsu Chae}.} \bibinfo{year}{2021}\natexlab{}.
\newblock \showarticletitle{{KoDF}: {A} {Large}-{Scale} {Korean} {DeepFake} {Detection} {Dataset}}. In \bibinfo{booktitle}{\emph{Proceedings of the {IEEE}/{CVF} {International} {Conference} on {Computer} {Vision}}}. \bibinfo{pages}{10744--10753}.
\newblock
\urldef\tempurl%
\url{https://openaccess.thecvf.com/content/ICCV2021/html/Kwon_KoDF_A_Large-Scale_Korean_DeepFake_Detection_Dataset_ICCV_2021_paper.html}
\showURL{%
\tempurl}


\bibitem[Li et~al\mbox{.}(2025)]%
        {liLatentSync2025}
\bibfield{author}{\bibinfo{person}{Chunyu Li}, \bibinfo{person}{Chao Zhang}, \bibinfo{person}{Weikai Xu}, \bibinfo{person}{Jingyu Lin}, \bibinfo{person}{Jinghui Xie}, \bibinfo{person}{Weiguo Feng}, \bibinfo{person}{Bingyue Peng}, \bibinfo{person}{Cunjian Chen}, {and} \bibinfo{person}{Weiwei Xing}.} \bibinfo{year}{2025}\natexlab{}.
\newblock \bibinfo{title}{{LatentSync}: {Taming} {Audio}-{Conditioned} {Latent} {Diffusion} {Models} for {Lip} {Sync} with {SyncNet} {Supervision}}.
\newblock
\href{https://doi.org/10.48550/arXiv.2412.09262}{doi:\nolinkurl{10.48550/arXiv.2412.09262}}
\newblock
\shownote{arXiv:2412.09262 [cs]}.


\bibitem[Li et~al\mbox{.}(2020)]%
        {liCelebDF2020}
\bibfield{author}{\bibinfo{person}{Yuezun Li}, \bibinfo{person}{Xin Yang}, \bibinfo{person}{Pu Sun}, \bibinfo{person}{Honggang Qi}, {and} \bibinfo{person}{Siwei Lyu}.} \bibinfo{year}{2020}\natexlab{}.
\newblock \showarticletitle{Celeb-{DF}: {A} {Large}-{Scale} {Challenging} {Dataset} for {DeepFake} {Forensics}}. In \bibinfo{booktitle}{\emph{Proceedings of the {IEEE}/{CVF} {Conference} on {Computer} {Vision} and {Pattern} {Recognition}}}. \bibinfo{pages}{3207--3216}.
\newblock
\urldef\tempurl%
\url{https://openaccess.thecvf.com/content_CVPR_2020/html/Li_Celeb-DF_A_Large-Scale_Challenging_Dataset_for_DeepFake_Forensics_CVPR_2020_paper.html}
\showURL{%
\tempurl}


\bibitem[Liu et~al\mbox{.}(2023)]%
        {liuASVspoof2023}
\bibfield{author}{\bibinfo{person}{Xuechen Liu}, \bibinfo{person}{Xin Wang}, \bibinfo{person}{Md Sahidullah}, \bibinfo{person}{Jose Patino}, \bibinfo{person}{Héctor Delgado}, \bibinfo{person}{Tomi Kinnunen}, \bibinfo{person}{Massimiliano Todisco}, \bibinfo{person}{Junichi Yamagishi}, \bibinfo{person}{Nicholas Evans}, \bibinfo{person}{Andreas Nautsch}, {and} \bibinfo{person}{Kong~Aik Lee}.} \bibinfo{year}{2023}\natexlab{}.
\newblock \showarticletitle{{ASVspoof} 2021: {Towards} {Spoofed} and {Deepfake} {Speech} {Detection} in the {Wild}}.
\newblock \bibinfo{journal}{\emph{IEEE/ACM Transactions on Audio, Speech, and Language Processing}}  \bibinfo{volume}{31} (\bibinfo{year}{2023}), \bibinfo{pages}{2507--2522}.
\newblock
\showISSN{2329-9304}
\href{https://doi.org/10.1109/TASLP.2023.3285283}{doi:\nolinkurl{10.1109/TASLP.2023.3285283}}


\bibitem[Mukhopadhyay et~al\mbox{.}(2024)]%
        {mukhopadhyayDiff2Lip2024}
\bibfield{author}{\bibinfo{person}{Soumik Mukhopadhyay}, \bibinfo{person}{Saksham Suri}, \bibinfo{person}{Ravi~Teja Gadde}, {and} \bibinfo{person}{Abhinav Shrivastava}.} \bibinfo{year}{2024}\natexlab{}.
\newblock \showarticletitle{{Diff2Lip}: {Audio} {Conditioned} {Diffusion} {Models} for {Lip}-{Synchronization}}. In \bibinfo{booktitle}{\emph{Proceedings of the {IEEE}/{CVF} {Winter} {Conference} on {Applications} of {Computer} {Vision}}}. \bibinfo{pages}{5292--5302}.
\newblock
\urldef\tempurl%
\url{https://openaccess.thecvf.com/content/WACV2024/html/Mukhopadhyay_Diff2Lip_Audio_Conditioned_Diffusion_Models_for_Lip-Synchronization_WACV_2024_paper.html}
\showURL{%
\tempurl}


\bibitem[Narang et~al\mbox{.}(2025)]%
        {narang2025laylens}
\bibfield{author}{\bibinfo{person}{Abhijeet Narang}, \bibinfo{person}{Parul Gupta}, \bibinfo{person}{Liuyijia Su}, {and} \bibinfo{person}{Abhinav Dhall}.} \bibinfo{year}{2025}\natexlab{}.
\newblock \showarticletitle{LayLens: Improving Deepfake Understanding through Simplified Explanations}.
\newblock \bibinfo{journal}{\emph{arXiv preprint arXiv:2507.10066}} (\bibinfo{year}{2025}).
\newblock


\bibitem[Narayan et~al\mbox{.}(2023)]%
        {narayanDFPlatter2023}
\bibfield{author}{\bibinfo{person}{Kartik Narayan}, \bibinfo{person}{Harsh Agarwal}, \bibinfo{person}{Kartik Thakral}, \bibinfo{person}{Surbhi Mittal}, \bibinfo{person}{Mayank Vatsa}, {and} \bibinfo{person}{Richa Singh}.} \bibinfo{year}{2023}\natexlab{}.
\newblock \showarticletitle{{DF}-{Platter}: {Multi}-{Face} {Heterogeneous} {Deepfake} {Dataset}}. In \bibinfo{booktitle}{\emph{Proceedings of the {IEEE}/{CVF} {Conference} on {Computer} {Vision} and {Pattern} {Recognition}}}. \bibinfo{pages}{9739--9748}.
\newblock
\urldef\tempurl%
\url{https://openaccess.thecvf.com/content/CVPR2023/html/Narayan_DF-Platter_Multi-Face_Heterogeneous_Deepfake_Dataset_CVPR_2023_paper.html}
\showURL{%
\tempurl}


\bibitem[Nick and Andrew(2019)]%
        {nickContributing2019}
\bibfield{author}{\bibinfo{person}{Dufou Nick} {and} \bibinfo{person}{Jigsaw Andrew}.} \bibinfo{year}{2019}\natexlab{}.
\newblock \bibinfo{title}{Contributing {Data} to {Deepfake} {Detection} {Research}}.
\newblock
\urldef\tempurl%
\url{http://ai.googleblog.com/2019/09/contributing-data-to-deepfake-detection.html}
\showURL{%
\tempurl}


\bibitem[{OpenAI}(2024)]%
        {openaiGPT4o2024a}
\bibfield{author}{\bibinfo{person}{{OpenAI}}.} \bibinfo{year}{2024}\natexlab{}.
\newblock \bibinfo{title}{{GPT}-4o {System} {Card}}.
\newblock
\urldef\tempurl%
\url{http://arxiv.org/abs/2410.21276}
\showURL{%
\tempurl}


\bibitem[Ouyang et~al\mbox{.}(2022)]%
        {ouyangTraining2022}
\bibfield{author}{\bibinfo{person}{Long Ouyang}, \bibinfo{person}{Jeffrey Wu}, \bibinfo{person}{Xu Jiang}, \bibinfo{person}{Diogo Almeida}, \bibinfo{person}{Carroll Wainwright}, \bibinfo{person}{Pamela Mishkin}, \bibinfo{person}{Chong Zhang}, \bibinfo{person}{Sandhini Agarwal}, \bibinfo{person}{Katarina Slama}, \bibinfo{person}{Alex Ray}, \bibinfo{person}{John Schulman}, \bibinfo{person}{Jacob Hilton}, \bibinfo{person}{Fraser Kelton}, \bibinfo{person}{Luke Miller}, \bibinfo{person}{Maddie Simens}, \bibinfo{person}{Amanda Askell}, \bibinfo{person}{Peter Welinder}, \bibinfo{person}{Paul~F. Christiano}, \bibinfo{person}{Jan Leike}, {and} \bibinfo{person}{Ryan Lowe}.} \bibinfo{year}{2022}\natexlab{}.
\newblock \showarticletitle{Training language models to follow instructions with human feedback}. In \bibinfo{booktitle}{\emph{Advances in {Neural} {Information} {Processing} {Systems}}}, Vol.~\bibinfo{volume}{35}. \bibinfo{publisher}{Curran Associates, Inc.}, \bibinfo{pages}{27730--27744}.
\newblock
\urldef\tempurl%
\url{https://proceedings.neurips.cc/paper_files/paper/2022/hash/b1efde53be364a73914f58805a001731-Abstract-Conference.html}
\showURL{%
\tempurl}


\bibitem[Pal et~al\mbox{.}(2024)]%
        {palSemiTruths2024}
\bibfield{author}{\bibinfo{person}{Anisha Pal}, \bibinfo{person}{Julia Kruk}, \bibinfo{person}{Mansi Phute}, \bibinfo{person}{Manognya Bhattaram}, \bibinfo{person}{Diyi Yang}, \bibinfo{person}{Duen~Horng Chau}, {and} \bibinfo{person}{Judy Hoffman}.} \bibinfo{year}{2024}\natexlab{}.
\newblock \showarticletitle{Semi-{Truths}: {A} {Large}-{Scale} {Dataset} of {AI}-{Augmented} {Images} for {Evaluating} {Robustness} of {AI}-{Generated} {Image} detectors}. In \bibinfo{booktitle}{\emph{Advances in {Neural} {Information} {Processing} {Systems}}}, Vol.~\bibinfo{volume}{37}. \bibinfo{pages}{118025--118051}.
\newblock
\urldef\tempurl%
\url{https://proceedings.neurips.cc/paper_files/paper/2024/hash/d5cdf7e56422f2a229c497dd89c3b995-Abstract-Datasets_and_Benchmarks_Track.html}
\showURL{%
\tempurl}


\bibitem[Prajwal et~al\mbox{.}(2020)]%
        {prajwalLip2020}
\bibfield{author}{\bibinfo{person}{K~R Prajwal}, \bibinfo{person}{Rudrabha Mukhopadhyay}, \bibinfo{person}{Vinay~P. Namboodiri}, {and} \bibinfo{person}{C.V. Jawahar}.} \bibinfo{year}{2020}\natexlab{}.
\newblock \showarticletitle{A {Lip} {Sync} {Expert} {Is} {All} {You} {Need} for {Speech} to {Lip} {Generation} {In} the {Wild}}. In \bibinfo{booktitle}{\emph{Proceedings of the 28th {ACM} {International} {Conference} on {Multimedia}}} \emph{(\bibinfo{series}{{MM} '20})}. \bibinfo{publisher}{Association for Computing Machinery}, \bibinfo{address}{New York, NY, USA}, \bibinfo{pages}{484--492}.
\newblock
\showISBNx{978-1-4503-7988-5}
\href{https://doi.org/10.1145/3394171.3413532}{doi:\nolinkurl{10.1145/3394171.3413532}}


\bibitem[Pérez-Vieites et~al\mbox{.}(2024)]%
        {perez-vieitesVigo2024}
\bibfield{author}{\bibinfo{person}{Diego Pérez-Vieites}, \bibinfo{person}{Juan~José Moreira-Pérez}, \bibinfo{person}{Ángel Aragón-Kifute}, \bibinfo{person}{Raquel Román-Sarmiento}, {and} \bibinfo{person}{Rubén Castro-González}.} \bibinfo{year}{2024}\natexlab{}.
\newblock \showarticletitle{Vigo: {Audiovisual} {Fake} {Detection} and {Segment} {Localization}}. In \bibinfo{booktitle}{\emph{Proceedings of the 32nd {ACM} {International} {Conference} on {Multimedia}}} \emph{(\bibinfo{series}{{MM} '24})}. \bibinfo{publisher}{Association for Computing Machinery}, \bibinfo{address}{New York, NY, USA}, \bibinfo{pages}{11360--11364}.
\newblock
\showISBNx{9798400706868}
\href{https://doi.org/10.1145/3664647.3688983}{doi:\nolinkurl{10.1145/3664647.3688983}}


\bibitem[Rossler et~al\mbox{.}(2019)]%
        {rosslerFaceForensics2019}
\bibfield{author}{\bibinfo{person}{Andreas Rossler}, \bibinfo{person}{Davide Cozzolino}, \bibinfo{person}{Luisa Verdoliva}, \bibinfo{person}{Christian Riess}, \bibinfo{person}{Justus Thies}, {and} \bibinfo{person}{Matthias Niessner}.} \bibinfo{year}{2019}\natexlab{}.
\newblock \showarticletitle{{FaceForensics}++: {Learning} to {Detect} {Manipulated} {Facial} {Images}}. In \bibinfo{booktitle}{\emph{Proceedings of the {IEEE}/{CVF} {International} {Conference} on {Computer} {Vision}}}. \bibinfo{pages}{1--11}.
\newblock
\urldef\tempurl%
\url{https://openaccess.thecvf.com/content_ICCV_2019/html/Rossler_FaceForensics_Learning_to_Detect_Manipulated_Facial_Images_ICCV_2019_paper.html}
\showURL{%
\tempurl}


\bibitem[Singh et~al\mbox{.}(2023)]%
        {singhHave2023}
\bibfield{author}{\bibinfo{person}{Monisha Singh}, \bibinfo{person}{Ximi Hoque}, \bibinfo{person}{Donghuo Zeng}, \bibinfo{person}{Yanan Wang}, \bibinfo{person}{Kazushi Ikeda}, {and} \bibinfo{person}{Abhinav Dhall}.} \bibinfo{year}{2023}\natexlab{}.
\newblock \showarticletitle{Do {I} {Have} {Your} {Attention}: {A} {Large} {Scale} {Engagement} {Prediction} {Dataset} and {Baselines}}. In \bibinfo{booktitle}{\emph{Proceedings of the 25th {International} {Conference} on {Multimodal} {Interaction}}} \emph{(\bibinfo{series}{{ICMI} '23})}. \bibinfo{publisher}{Association for Computing Machinery}, \bibinfo{address}{New York, NY, USA}, \bibinfo{pages}{174--182}.
\newblock
\showISBNx{9798400700552}
\href{https://doi.org/10.1145/3577190.3614164}{doi:\nolinkurl{10.1145/3577190.3614164}}


\bibitem[Thakral et~al\mbox{.}(2024)]%
        {thakralILLUSION2024}
\bibfield{author}{\bibinfo{person}{Kartik Thakral}, \bibinfo{person}{Rishabh Ranjan}, \bibinfo{person}{Akanksha Singh}, \bibinfo{person}{Akshat Jain}, \bibinfo{person}{Mayank Vatsa}, {and} \bibinfo{person}{Richa Singh}.} \bibinfo{year}{2024}\natexlab{}.
\newblock \showarticletitle{{ILLUSION}: {Unveiling} {Truth} with a {Comprehensive} {Multi}-{Modal}, {Multi}-{Lingual} {Deepfake} {Dataset}}. In \bibinfo{booktitle}{\emph{The {Thirteenth} {International} {Conference} on {Learning} {Representations}}}.
\newblock
\urldef\tempurl%
\url{https://openreview.net/forum?id=qnlG3zPQUy}
\showURL{%
\tempurl}


\bibitem[Wang et~al\mbox{.}(2023)]%
        {wangSeeing2023}
\bibfield{author}{\bibinfo{person}{Jiadong Wang}, \bibinfo{person}{Xinyuan Qian}, \bibinfo{person}{Malu Zhang}, \bibinfo{person}{Robby~T. Tan}, {and} \bibinfo{person}{Haizhou Li}.} \bibinfo{year}{2023}\natexlab{}.
\newblock \showarticletitle{Seeing {What} {You} {Said}: {Talking} {Face} {Generation} {Guided} by a {Lip} {Reading} {Expert}}. In \bibinfo{booktitle}{\emph{Proceedings of the {IEEE}/{CVF} {Conference} on {Computer} {Vision} and {Pattern} {Recognition}}}. \bibinfo{pages}{14653--14662}.
\newblock
\urldef\tempurl%
\url{https://openaccess.thecvf.com/content/CVPR2023/html/Wang_Seeing_What_You_Said_Talking_Face_Generation_Guided_by_a_CVPR_2023_paper.html}
\showURL{%
\tempurl}


\bibitem[Wang et~al\mbox{.}(2024)]%
        {wangBuilding2024}
\bibfield{author}{\bibinfo{person}{Yifan Wang}, \bibinfo{person}{Xuecheng Wu}, \bibinfo{person}{Jia Zhang}, \bibinfo{person}{Mohan Jing}, \bibinfo{person}{Keda Lu}, \bibinfo{person}{Jun Yu}, \bibinfo{person}{Wen Su}, \bibinfo{person}{Fang Gao}, \bibinfo{person}{Qingsong Liu}, \bibinfo{person}{Jianqing Sun}, {and} \bibinfo{person}{Jiaen Liang}.} \bibinfo{year}{2024}\natexlab{}.
\newblock \showarticletitle{Building {Robust} {Video}-{Level} {Deepfake} {Detection} via {Audio}-{Visual} {Local}-{Global} {Interactions}}. In \bibinfo{booktitle}{\emph{Proceedings of the 32nd {ACM} {International} {Conference} on {Multimedia}}} \emph{(\bibinfo{series}{{MM} '24})}. \bibinfo{publisher}{Association for Computing Machinery}, \bibinfo{address}{New York, NY, USA}, \bibinfo{pages}{11370--11376}.
\newblock
\showISBNx{9798400706868}
\href{https://doi.org/10.1145/3664647.3688985}{doi:\nolinkurl{10.1145/3664647.3688985}}


\bibitem[Yang et~al\mbox{.}(2019)]%
        {yangExposing2019}
\bibfield{author}{\bibinfo{person}{Xin Yang}, \bibinfo{person}{Yuezun Li}, {and} \bibinfo{person}{Siwei Lyu}.} \bibinfo{year}{2019}\natexlab{}.
\newblock \showarticletitle{Exposing {Deep} {Fakes} {Using} {Inconsistent} {Head} {Poses}}. In \bibinfo{booktitle}{\emph{{IEEE} {International} {Conference} on {Acoustics}, {Speech} and {Signal} {Processing} ({ICASSP})}}. \bibinfo{pages}{8261--8265}.
\newblock
\href{https://doi.org/10.1109/ICASSP.2019.8683164}{doi:\nolinkurl{10.1109/ICASSP.2019.8683164}}
\newblock
\shownote{ISSN: 2379-190X}.


\bibitem[Zhang et~al\mbox{.}(2024)]%
        {zhangMFMS2024}
\bibfield{author}{\bibinfo{person}{Yi Zhang}, \bibinfo{person}{Changtao Miao}, \bibinfo{person}{Man Luo}, \bibinfo{person}{Jianshu Li}, \bibinfo{person}{Wenzhong Deng}, \bibinfo{person}{Weibin Yao}, \bibinfo{person}{Zhe Li}, \bibinfo{person}{Bingyu Hu}, \bibinfo{person}{Weiwei Feng}, \bibinfo{person}{Tao Gong}, {and} \bibinfo{person}{Qi Chu}.} \bibinfo{year}{2024}\natexlab{}.
\newblock \showarticletitle{{MFMS}: {Learning} {Modality}-{Fused} and {Modality}-{Specific} {Features} for {Deepfake} {Detection} and {Localization} {Tasks}}. In \bibinfo{booktitle}{\emph{Proceedings of the 32nd {ACM} {International} {Conference} on {Multimedia}}} \emph{(\bibinfo{series}{{MM} '24})}. \bibinfo{publisher}{Association for Computing Machinery}, \bibinfo{address}{New York, NY, USA}, \bibinfo{pages}{11365--11369}.
\newblock
\showISBNx{9798400706868}
\href{https://doi.org/10.1145/3664647.3688984}{doi:\nolinkurl{10.1145/3664647.3688984}}


\bibitem[Zhou et~al\mbox{.}(2021)]%
        {zhouFace2021}
\bibfield{author}{\bibinfo{person}{Tianfei Zhou}, \bibinfo{person}{Wenguan Wang}, \bibinfo{person}{Zhiyuan Liang}, {and} \bibinfo{person}{Jianbing Shen}.} \bibinfo{year}{2021}\natexlab{}.
\newblock \showarticletitle{Face {Forensics} in the {Wild}}. In \bibinfo{booktitle}{\emph{Proceedings of the {IEEE}/{CVF} {Conference} on {Computer} {Vision} and {Pattern} {Recognition}}}. \bibinfo{pages}{5778--5788}.
\newblock
\urldef\tempurl%
\url{https://openaccess.thecvf.com/content/CVPR2021/html/Zhou_Face_Forensics_in_the_Wild_CVPR_2021_paper.html}
\showURL{%
\tempurl}


\bibitem[Zi et~al\mbox{.}(2020)]%
        {ziWildDeepfake2020}
\bibfield{author}{\bibinfo{person}{Bojia Zi}, \bibinfo{person}{Minghao Chang}, \bibinfo{person}{Jingjing Chen}, \bibinfo{person}{Xingjun Ma}, {and} \bibinfo{person}{Yu-Gang Jiang}.} \bibinfo{year}{2020}\natexlab{}.
\newblock \showarticletitle{{WildDeepfake}: {A} {Challenging} {Real}-{World} {Dataset} for {Deepfake} {Detection}}. In \bibinfo{booktitle}{\emph{Proceedings of the 28th {ACM} {International} {Conference} on {Multimedia}}} \emph{(\bibinfo{series}{{MM} '20})}. \bibinfo{publisher}{Association for Computing Machinery}, \bibinfo{address}{New York, NY, USA}, \bibinfo{pages}{2382--2390}.
\newblock
\showISBNx{978-1-4503-7988-5}
\href{https://doi.org/10.1145/3394171.3413769}{doi:\nolinkurl{10.1145/3394171.3413769}}


\bibitem[Zingarini et~al\mbox{.}(2024)]%
        {zingariniM3DSYNTH2024}
\bibfield{author}{\bibinfo{person}{G. Zingarini}, \bibinfo{person}{D. Cozzolino}, \bibinfo{person}{R. Corvi}, \bibinfo{person}{G. Poggi}, {and} \bibinfo{person}{L. Verdoliva}.} \bibinfo{year}{2024}\natexlab{}.
\newblock \showarticletitle{{M3DSYNTH}: {A} {Dataset} of {Medical} {3D} {Images} with {AI}-{Generated} {Local} {Manipulations}}. In \bibinfo{booktitle}{\emph{{ICASSP} 2024 - 2024 {IEEE} {International} {Conference} on {Acoustics}, {Speech} and {Signal} {Processing} ({ICASSP})}}. \bibinfo{pages}{13176--13180}.
\newblock
\href{https://doi.org/10.1109/ICASSP48485.2024.10446605}{doi:\nolinkurl{10.1109/ICASSP48485.2024.10446605}}
\newblock
\shownote{ISSN: 2379-190X}.


\end{thebibliography}

\end{document}